\documentclass{article}

\usepackage[nonatbib,final]{nips2016} % produce camera-ready copy

\usepackage{amsmath,amssymb}
\usepackage{graphicx,subfigure}
\title{A Powerful Generative Model Using Random Weights for the Deep Image Representation}

\author{
  Kun He\thanks{
  The three authors contribute equally. } \\
  School of Computer Science and Technology, \\
  Huazhong University of Science and Technology,
  Wuhan 430074, China;\\
  Department of Computer Science, Cornell University, Ithaca 14850, NY, USA  \\
  \texttt{brooklet60@hust.edu.cn, kh555@cs.cornell.edu} \\
  \AND
  Yan Wang \thanks{Corresponding author.}\\
 School of Computer Science and Technology, \\
 Huazhong University of Science and Technology,
 Wuhan 430074, China \\
 \texttt{yanwang@hust.edu.cn} \\
 \AND
  John Hopcroft \\
  Department of Computer Science,
  Cornell University, Ithaca 14850, NY, USA \\
  \texttt{jeh@cs.cornell.edu} \\
}

\begin{document}
% \nipsfinalcopy is no longer used

\maketitle

\begin{abstract}
To what extent is the success of deep visualization due to the training? Could we do deep visualization using untrained, random weight networks? To address this issue, we explore new and powerful generative models for three popular deep visualization tasks using untrained, random weight convolutional neural networks. First we invert representations in feature spaces and reconstruct images from white noise inputs. The reconstruction quality is statistically higher than that of the same method applied on well trained networks with the same architecture. Next we synthesize textures using scaled correlations of representations in multiple layers and our results are almost indistinguishable with the original natural texture and the synthesized textures based on the trained network. Third, by recasting the content of an image in the style of various artworks, we create artistic images with high perceptual quality, highly competitive to the prior work of Gatys et al. on pretrained networks. To our knowledge this is the first demonstration of image representations using untrained deep neural networks. Our work provides a new and fascinating tool to study the representation of deep network architecture and sheds light on new understandings on deep visualization.
\end{abstract}

\section{Introduction}
\label{sec:introduction}

In recent years, Deep Neural Networks (DNNs), especially Convolutional Neural Networks (CNNs), have demonstrated highly competitive results on object recognition and image classification~\cite{ILSVRC,Alex2012NIPS,Simonyan14VGG,He2015Residual}.
With advances in training,  there is a growing trend towards %understanding how DNNs works, and how computations DNNs perform at intermediate layers.
understanding the inner working of these deep networks.
By training on a very large image data set, DNNs develop a representation of images that makes object information increasingly explicit at various levels of the hierarchical architecture.
Significant visualization techniques have been developed to understand the deep image representations on trained networks~\cite{Erhan2009, Simonyan13Visualise,Mahendran2015CVPR, Gatys2015texture, yosinski2015ICML,Dosovitskiy2015inverting,nguyen2015deepFool}.

Inversion techniques have been developed to create synthetic images with feature representations similar to the representations of an original image in one or several layers of the network.
Feature representations are a function $\Phi$ of the source image $\texttt{x}_0.$ An approximate inverse $\Phi^{-1}$ is used to construct a new image $\texttt{x}$ from the code $\Phi(\texttt{x}_0)$ by reducing some statistical discrepancy between $\Phi(\texttt{x})$ and $\Phi(\texttt{x}_0)$.
Mahendran et al. \cite{Mahendran2015CVPR} use the pretrained CNN AlexNet~\cite{Alex2012NIPS} and define a squared Euclidean loss on the activations to capture the representation differences and reconstruct the image.
Gatys et al.~\cite{Gatys2015texture, Gatys2015texture2} define a squared loss on the correlations between feature maps of some layers and synthesize natural textures of high perceptual quality using the pretrained CNN called VGG~\cite{Simonyan14VGG}. Gatys et al.~\cite{Gatys2015Style} then combine the loss on the correlations as a proxy to the style of a painting and the loss on the activations to represent the content of an image, and successfully create artistic images by converting the artistic style to the content image,
inspiring several followups~\cite{nikulin2016style,Johnson2016perceptual}.
Another stream of visualization aims to understand what each neuron has learned in a pretrained network and synthesize an image that maximally activates individual features~\cite{Erhan2009, yosinski2015ICML} or the class prediction scores~\cite{Simonyan13Visualise}. Nguyen et al. further try multifaceted visualization to separate and visualize different features that a neuron learns~\cite{Jason2016Multifaceted}.

Feature inversion and neural activation maximization both start from a white noise image and calculate the gradient via backpropagation to morph the white noise image and output a natural image.
In addition, some regularizers are incorporated as a natural image prior to improve the visualization quality, including $\alpha-$norm~\cite{Simonyan13Visualise}, total variation\cite{Mahendran2015CVPR}, jitter\cite{Mahendran2015CVPR}, Gaussian blur~\cite{yosinski2015ICML}, data-driven patch priors~\cite{Wei2015Understanding}, etc.
The method of visualizing the feature representation on the intermediate layers sheds light on the information represented at each layer of the pretrained CNN.

A third set of researchers  trains a separate feed-forward CNN with upconvolutional layers using representations or correlations of the feature maps produced in the original network as the input and the source image as the target to learn the inversion of the original network. The philosophy is to train another neural network to inverse the representation and speedup the visualization on image reconstruction\cite{Dosovitskiy2015inverting, Dosovitskiy2016generating}, texture synthesis\cite{Ulyanov2016texture} or even style transfer\cite{Johnson2016perceptual}.
Instead of designing a natural prior, some researchers incorporate adversarial training\cite{Goodfellow2014NIPS} to improve the realism of the generated images\cite{Dosovitskiy2016generating}.
Their trained upconvolutional network could give similar qualitative results as the inversion technique does and is two or three orders of magnitude faster, as the previous inversion technique needs a forward and backward pass through the pretrained network.
This technique is slightly different from the previous two in that it does not focus on understanding the original CNN but on the visualization task.

It is well recognized that deep visualization techniques conduct a direct analysis of the visual information contained in image representations, and help us understand the representation encoded at the intermediate layers of the well trained DNNs.
In this paper, we raise a fundamental issue that other researchers rarely address:  \textbf{Could we do deep visualization using untrained, random weight DNNs? This would allow us to separate the contribution of training from the contribution of the network structure.
It might even give us a method to evaluate deep network architectures without spending days and significant computing resources in training networks so that we could compare them.}
Though Gray et al. demonstrated that the VGG architecture with random weights failed in generating textures and resulted in white noise images in an experiment indicating the trained filters might be crucial for texture generation~\cite{Gatys2015texture},
we conjecture the success of deep visualization mainly originates from the intrinsic nonlinearity and complexity of the deep network hierarchical structure rather than from the training, and that the architecture itself may cause the inversion invariant to the original image.
Gatys et al.'s unsuccessful attempt on the texture synthesis using the VGG architecture with random weights may be due to their inappropriate scale of the weighting factors.

To verify our hypothesis, we try three popular inversion tasks for visualization using the CNN architecture with random weights. Our results strongly suggest that this is true.
Applying inversion techniques on the \emph{untrained} VGG with random weights, we
 reconstruct high perceptual quality images. The results are qualitatively better than the reconstructed images produced on the pretrained VGG with the same architecture.
 %Therefore, what is gained by the training and what is gained by just the CNN network structure?
%And what else could we explore using networks with purely random weights?
Then, we try to synthesize natural textures using the random weight VGG. With automatic normalization to scale the squared correlation loss for different activation layers, we succeed in generating similar textures as the prior work of Gatys et al.~\cite{Gatys2015texture} on well-trained VGG.
Furthermore, we continue the experiments on style transfer, combining the content of an image and the style of an artwork,
and create artistic imagery using random weight CNN.

 %We wish our work shed light on what is learned by the training and what is learned by purely the network structure.
To our knowledge this is the first demonstration of image representations using untrained deep neural networks.
 Our work  provides a new and fascinating tool to study the perception and representation of deep network architecture, and
shed light on new understandings on deep visualization. % and gives a way to explore the effects due to training and the effects due to the network structure.
 Our work will inspire more possibilities of using the generative power of CNNs with random weights, which do not need long training time on multi-GPUs.
 Furthermore, it is very hard to prove why trained deep neural networks work so well.
 Based on the networks with random weights, we might be able to prove some properties of the deep networks.  Our work using random weights shows a possible way to start developing a theory of deep learning since with well-trained weights, theorems might be impossible.

%representation inversion
\vspace{-0.5em}
\section{Convolutional neural network}
\label{sec:network}
%\input{020Network}

\begin{figure}[htbp]
\centering
\vspace{-0.5em}
\includegraphics[width=5in]{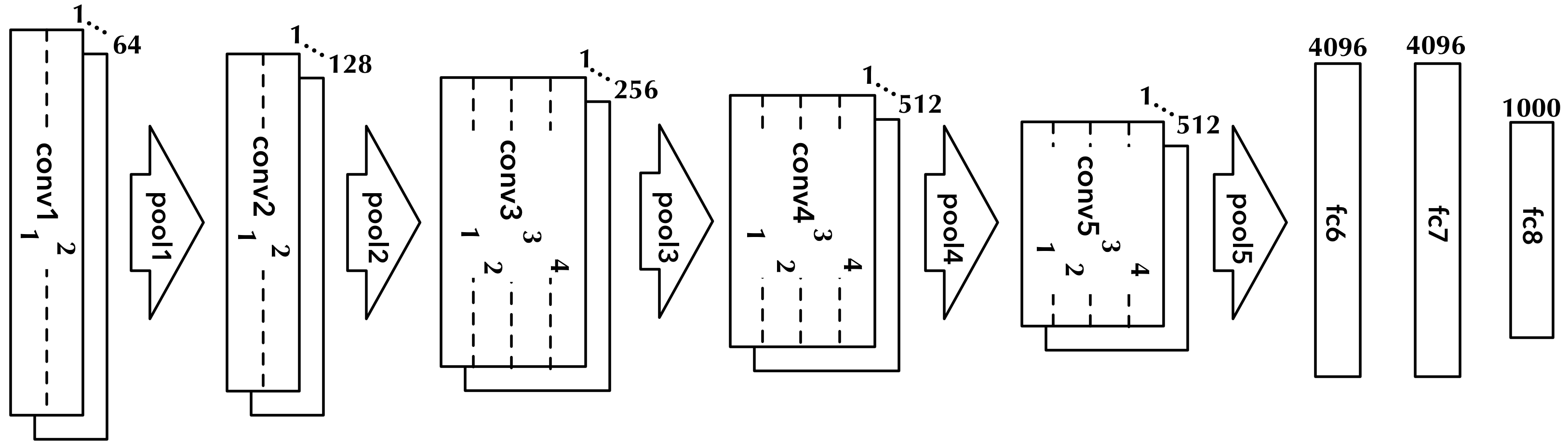}
\vspace{-1em}
\caption{Architecture of the VGG network.}
\vspace{-0.5em}
\label{Fig:VGG}
\end{figure}

%\makeindex book.idx \makeindex book.idx
VGG-19~\cite{Simonyan14VGG} is a convolutional neural network trained on the 1.3 million-image ILSVRC 2012 ImageNet dataset~\cite{ILSVRC} using the Caffe-framework~\cite{jia2014caffe}.
The VGG architecture has 16 convolutional and 5 pooling layers, followed by 3 fully connected layers as illustrated in Figure \ref{Fig:VGG}. The filters are of size $3 \times 3 \times N_l$  where $N_l$ is the number of feature maps (or channels). Pooling is applied between the convolutional layers with $2 \times 2$ spatial down-sampling on the feature maps.
A pre-processing step subtracts the mean RGB value of each pixel of the training set.% so that the mean value is 0.

Gatys et al. re-train the VGG-19 network using average pooling instead of maximum pooling, which they suggest could improve the gradient flow and obtain slightly better results~\cite{Gatys2015texture}.
 They only consider the convolutional and pooling layers for texture synthesis, and they rescale the weights such that the mean activation of each filter over the images and positions is 1. Their trained network is denoted as \emph{VGG} in the following discussion.

We adopt the same architecture, replacing the weights with purely random values from a Gaussian distribution $N(0, \sigma)$. The standard deviation, $\sigma,$ is set to a small number like 0.015 in the experiments.
The VGG-based random weight network created as described in the method section is used as our reference network, denoted as \emph{ranVGG} in the following discussion.

\vspace{-0.5em}
\section{Methods}
\label{sec:methods}

In order to better understand the deep representation in the CNN architecture,
we focus on three tasks: inverting the image representation, synthesizing  texture, and creating artistic style images.
Our methods are similar in spirit to existing methods~\cite{Mahendran2015CVPR, Gatys2015texture,Gatys2015Style}.
The main difference is that we use random weights instead of trained weights, and we apply weighting factors determined by a pre-process to normalize the different impact scales of different activation layers on the input layer.
Another change is that we apply a greedy approach to build a ``stacked" random weight network using the inversion technique to stabilize the visualization quality.
%\footnote{The code can be downloaded from the GitHub after the review.}.
%based on the VGG-19 network architecture.

%\subsection{Inverting the deep representations}
\paragraph{Inverting deep representations.}
Given a representation function $F^l: \mathbb{R}^{H \times W \times C} \rightarrow \mathbb{R}^{N_l \times M_l}$ for the $l^{th}$ layer of a deep network and %a representation
 $F^l (\texttt{x}_0)$ for an input image $\texttt{x}_0$,
we want to reconstruct an image $\texttt{x}$ that minimizes the $L_2$ loss among the representations of $\texttt{x}_0$ and $\texttt{x}$. %squared Euclidean distance
\setlength{\belowdisplayskip}{2pt}
\setlength{\abovedisplayskip}{2pt}
\begin{equation}
\texttt{x}^* = \operatornamewithlimits{argmin}_{ \texttt{x} \in \mathbb{R}^{H \times W \times C} } \mathcal{L}_{content} (\texttt{x}, \texttt{x}_0, l) =  \operatornamewithlimits{argmin}_{ \texttt{x} \in \mathbb{R}^{H \times W \times C} } \frac{\omega_l}{2 N_l M_l} \lVert F^l(\texttt{x})- F^l(\texttt{x}_0) \rVert _2^2
%\texttt{x}^* = \operatornamewithlimits{argmin}_{ \texttt{x} \in \mathbb{R}^{H \times W \times C} } \mathcal{L} (F^l(\texttt{x}), F^l(\texttt{x}_0)) = \frac{1}{2 N_l \cdot M_l} \operatornamewithlimits{argmin}_{ \texttt{x} \in \mathbb{R}^{H \times W \times C} } \frac{\lVert F^l(\texttt{x})- F^l(\texttt{x}_0) \rVert _2^2}{  avg[F^l(\texttt{x}_0)] ^2 }
\label{eq:contentLoss}
\end{equation}

Here $H$ and $W$ denote the size of the image, $C = 3$ the color channels, and
 $\omega_l$ the weighting factor.
We regard the feature map matrix $F^l$ as the representation function of the $l^{th}$ layer
 which has $N_l \times M_l$ dimensions where $N_l$ is the number of distinct feature maps, each of size $M_l$ when vectorised.
$F_{ik}^l$ denotes the activation of the $i^{th}$ filter at position $k$.
%is described in the following by Equation (\ref{eq:EdisWl2}).

%The normalization by $\lVert F^l(\texttt{x}_0) \rVert _2^2$ doesn't affect the reconstruction result, but is added for the combination of content and style
%artwork generation described in the third portion of this section.
%And we normalize the objective by the square of the mean on elements in $F^l$.
The representations are a chain of non-linear filter banks even if untrained random weights are applied to the network.
We initialize the pre\_image with white noise,
%by minimizing the average squared Euclidean loss between the target image and the pre\_image,
and apply the L\_BFGS gradient descent using standard error backpropagation to morph the input pre\_image to the target.
 %\texttt{x}_{t+1}  = \texttt{x}_t - \frac{ \partial  \mathcal{L}^l(\texttt{x},\texttt{x}_0) }{ \partial \texttt{x} }
\setlength{\belowdisplayskip}{2pt}
\setlength{\abovedisplayskip}{2pt}
\begin{align}
 \texttt{x}_{t+1}  &= \texttt{x}_t - \left(\frac{ \partial  \mathcal{L}(\texttt{x},\texttt{x}_0,l) }{ \partial F^l}  \frac{\partial F^l}{\partial \texttt{x}}\right) \bigg|_{\texttt{x}_t}  \\
  \frac{\partial \mathcal{L}(\texttt{x},\texttt{x}_0,l)}{\partial{F^l_{i,k}}}\bigg|_{\texttt{x}_t}  &= \frac{1}{N^l M^l}(F^l(\texttt{x}_t)-F^l(\texttt{x}_0))_{i,k}  %\cdot
\end{align}

The weighting factor $\omega_{l}$ is applied to normalize the gradient impact on the morphing image $\texttt{x}$. We use a pre-processing procedure to determine the value of $\omega_{l}$.
%For the current representation layer $l$, we calculate the approximate maximum gradient by Equation (\ref{eq:EdisWl2}), and back propagate the gradient to the input layer,
%and use the mean gradient over all pixels and RGB channels as the weighting factor value.
For the current layer $l$, we approximately calculate the maximum possible gradient by Equation (\ref{eq:EdisWl2}), % the average absolute gradient over all pixels and color channels,
and back propagate the gradient to the input layer.
%The intuition on the estimation is that when $\texttt{x}$ approaches $\texttt{x}_0$, we could estimate the maximal gradient by replacing $\texttt{x}$ to $\texttt{x}_0$, and assuming the activations of the other image $\texttt{x}'$ are all zeros.
Then we regard the reciprocal of the absolute mean gradient over all pixels and RGB channels as the value of $\omega_{l}$ such that the gradient impact of different layers is approximately of the same scale.
This normalization doesn't affect the reconstruction from the activations of a single layer,
but is added for the combination of content and style for the style transfer task.
%And we normalize the objective by the square of the mean on elements in $F^l$.
\setlength{\belowdisplayskip}{2pt}
\setlength{\abovedisplayskip}{2pt}
%\begin{equation}
%\frac{\partial \mathcal{L}^l(\texttt{x}_0,\texttt{x}')}{\partial{F_{i,k}}}\bigg|_{F^l(\texttt{x}')=0} = \frac{1}{N^l M^l}F^l(\texttt{x}_0)_{i,k}
%\label{eq:EdisWl1}
%\end{equation}
\begin{equation}
\frac{1}{\omega_{l}} =\frac{1}{W H C}\bigg|  \displaystyle{ \sum_{i=1}^{W}} \displaystyle{ \sum_{j=1}^{H}} \displaystyle{\sum_{k=1}^{C}}\frac{\partial \mathcal{L}(\texttt{x}_0,\texttt{x}',l)}{\partial{\texttt{x}_{i,j,k}}} \bigg|_{F^l(\texttt{x}')=0}
\label{eq:EdisWl2}
\end{equation}

%Due to random weights, it's difficult to balance the different layer's influence
%on the reconstruct image. Therefore, we propose a pre-processing method to solve this problem.
%In every ? layer, we calculate the gradient on this layer and back propagate gradient from
% this layer to input layer. Then  we obtain the gradient on input layer, and calculate the mean,
% as $\omega_{l}$. When we do the inversing image or texture synthesis, the loss on $l$ layer needs
% to divide $\omega_{l}$. We want scale each layer to make sure that they have the same mean of gradient
% on input layer.

To stabilize the reconstruction quality, we apply a greedy approach to build a ``stacked" random weight network \emph{ranVGG} based on the VGG-19 architecture.
Select one single image as the reference image and starting from the first convolutional layer,
we build the stacked random weight VGG by sampling, selecting and fixing the weights of each layer in forward order.
For the current layer $l$, fix the weights of the previous $l-1$ layers and sample several sets of random weights connecting the $l^{th}$ layer. Then reconstruct the target image using the rectified representation of layer $l$, and choose weights yielding the smallest loss.
Experiments in the next section show our success on the reconstruction by using the untrained, random weight CNN, \emph{ranVGG}.

%\subsection{Texture synthesis}
\paragraph{Texture synthesis.} Can we synthesize natural textures based on the feature space of an untrained deep network?
To address this issue, we refer to the method proposed by Gatys et al.\cite{Gatys2015texture} and use the correlations between feature responses on each layer as the texture representation. %For layer $l$, we use the feature map matrix $F^l \in  \mathbb{R}^{N_l \times M_l}$  where
% $F_{ik}^l$ is the activation of the $i^{th}$ filter at position $k$.
The inner product between pairwise feature maps $i$ and $j$ within each layer $l$, $G_{ij}^l = \sum_{k} F_{ik}^lF_{jk}^l$, defines a gram matrix $G^l = F^l (F^l)^T$. % \in \mathbb{R}^{N_l \times N_l}$.
We seek a texture image $\texttt{x}$ that minimizes the $L_2$ loss among the correlations of the representations of several candidate layers for $\texttt{x}$ and a groundtruth image $\texttt{x}_0$.
%And a combination of the loss contributions of several candidate layers gives rise to total loss.
\setlength{\belowdisplayskip}{2pt}
\setlength{\abovedisplayskip}{2pt}
\begin{equation}
\texttt{x}^* = \operatornamewithlimits{argmin}_{ \texttt{x} \in \mathbb{R}^{H \times W \times C} } \mathcal{L}_{texture}  = \operatornamewithlimits{argmin}_{ \texttt{x} \in \mathbb{R}^{H \times W \times C} } \sum_{l \in L} \mu_l E(\texttt{x},\texttt{x}_0,l),
%\texttt{x}^* = \operatornamewithlimits{argmin}_{ \texttt{x} \in \mathbb{R}^{H \times W \times C} } \mathcal{L} [G^l(F^l(\texttt{x})), G^l(F^l(\texttt{x}_0))] = \frac{\mu_l}{4N_l^2 \cdot M_l^2} \operatornamewithlimits{argmin}_{ \texttt{x} \in \mathbb{R}^{H \times W \times C} }  \lVert G^l(F^l(\texttt{x}))- G^l(F^l(\texttt{x}_0)) \rVert _2^2
\label{eq:textureLoss}
\end{equation}
%\begin{equation}
%\texttt{x}^* = \operatornamewithlimits{argmin}_{ \texttt{x} \in \mathbb{R}^{H \times W \times C} } \mathcal{L} [G^l(F^l(\texttt{x})), G^l(F^l(\texttt{x}_0))] = \frac{1}{4N_l^2 \cdot M_l^2} \operatornamewithlimits{argmin}_{ \texttt{x} \in \mathbb{R}^{H \times W \times C} } \frac{\lVert G^l(F^l(\texttt{x}))- G^l(F^l(\texttt{x}_0)) \rVert _2^2}{ avg[ G^l(F^l(\texttt{x}_0)) ]^2}
%\label{eq:textureLoss}
%\end{equation}
where the contribution of layer $l$ to the total loss is defined as
\setlength{\belowdisplayskip}{2pt}
\setlength{\abovedisplayskip}{2pt}
\begin{equation}
E(\texttt{x},\texttt{x}_0,l)  = \frac{1}{4N_l^2 M_l^2}  \lVert G^l(F^l(\texttt{x}))- G^l(F^l(\texttt{x}_0)) \rVert _2^2.
\label{eq:textureLoss}
\end{equation}

The derivative of $E(\texttt{x},\texttt{x}_0,l)$ with respect to the activations $F^l$ in layer $l$ is~\cite{Gatys2015texture}:
\setlength{\belowdisplayskip}{2pt}
\setlength{\abovedisplayskip}{2pt}
\begin{equation}
\frac{\partial  E(\texttt{x},\texttt{x}_0,l)}{\partial{F^l_{i,k}}} = \frac{1}{N^2_l M^2_l}\{(F^l(\texttt{x}))^T [G^l(F^l(\texttt{x}))- G^l(F^l(\texttt{x}_0))]\}_{i,k}
\label{eq:textureLoss}
\end{equation}

%Due to random weights, it's difficult to balance the different layer's influence
%on the reconstruct image. Therefore, we propose a pre-processing method to solve this problem.
%In every ? layer, we calculate the gradient on this layer and back propagate gradient from
% this layer to input layer. Then  we obtain the gradient on input layer, and calculate the mean,
% as $\gamma^{l}$. When we do the inversing image or texture synthesis, the loss on $l$ layer needs
% to divide $\gamma^{l}$. We want scale each layer to make sure that they have the same mean of gradient
% on input layer.

%The weighting factor $\mu_{l}$ is applied to normalize the gradient impact on the input layer.
%For each candidate layer $l$, we approximately calculate the gradient scale by Equation (\ref{eq:TextureW})
%and back propagate the gradient to the input layer.
%Then we regard the reciprocal of the mean gradient over all pixels and RGB channels as the value of $\mu_{l}$.

The weighting factor $\mu_{l}$ is defined similarly to $\omega_{l}$,
 but here we use the loss contribution $E(\texttt{x},\texttt{x}_0,l)$ of layer $l$ to get its gradient impact on the input layer.
\setlength{\belowdisplayskip}{2pt}
\setlength{\abovedisplayskip}{2pt}
%\begin{equation}
% \frac{\partial E(\texttt{x},\texttt{x}_0,l)}{\partial{F^l_{i,k}}} = \frac{1}{N^2_l M^2_l}[(F^l(\texttt{x}_0))^T(G^l(F^l(\texttt{x}_0))-0)]_{i,k}
%\end{equation}
\begin{equation}
\frac{1}{\mu_{l}} =\frac{1}{W H C}\bigg|  \displaystyle{ \sum_{i}^{W}} \displaystyle{ \sum_{j}^{H}} \displaystyle{\sum_{k}^{C}}\frac{\partial E(\texttt{x}_0,\texttt{x}', l)}{\partial{\texttt{x}_{i,j,k}}} \bigg|_{F^l(\texttt{x}')=0}
\label{eq:TextureW}
\end{equation}

We then perform the L\_BFGS gradient descent using standard error backpropagation to morph the input image to a synthesized texture image using the untrained \emph{ranVGG}.

\paragraph{Style transfer.} Can we use the untrained deep network to create artistic images? Referring to the prior work of Gatys et al.\cite{Gatys2015Style} from the feature responses of VGG trained on ImageNet, we use an untrained VGG and succeed in separating and recombining content and style of arbitrary images.
The objective requires terms for content and style respectively with suitable combination factors.
For content we use the method of reconstruction on medium layer representations, and for style we use the method of synthesising texture on some lower through higher layer representation correlations.

%by combining the reconstruction loss $\mathcal{L}_{content}$ and the texture loss $\mathcal{L}_{texture}$.

Let $\texttt{x}_c$ be the content image and $\texttt{x}_s$ the style image. We combine the content of the former and the style of the latter by optimizing the following objective:
\setlength{\belowdisplayskip}{2pt}
\setlength{\abovedisplayskip}{2pt}
\begin{equation}
\texttt{x}^* = \operatornamewithlimits{argmin}_{ \texttt{x} \in \mathbb{R}^{H \times W \times C} }
\alpha \mathcal{L}_{content}(\texttt{x}, \texttt{x}_c) + \beta \mathcal{L}_{texture}(\texttt{x}, \texttt{x}_s) + \gamma \mathcal{R} (\texttt{x})
\label{eq:artLoss}
\end{equation}

Here $\alpha$ and $\beta$ are the contributing factors for content and style respectively. We apply a regularizer $\mathcal{R} (\texttt{x})$, total variation(TV)~\cite{Mahendran2015CVPR} defined as the squared sum on the adjacent pixel's difference of $\texttt{x}$, to encourage the spatial smoothness in the output image.
%In the experiments, we select conv2\_2 as the content layer, and use the combination of conv1\_1,  conv2\_1, ..., conv5\_1 as the style. We set the ratio of $\alpha :\beta : \gamma = 100 : 1 : 1000$.

\vspace{-0.5em}
\section{Experiments}

This section evaluates the representation inversion, texture synthesis and style transfer results obtained by our model using the untrained network \emph{ranVGG}.
The input image is required to be of size $256 \times 256$ if we want to invert the representation of the fully connected layers. Else, the input could be of arbitrary size.
%For texture synthesis and style transfer which only need representations of the convolutional layers and pooling layers,

\label{sec:experiments}
%\input{040Experiments}
%Gatys et al.\cite{Gatys2015texture} have tried the VGG architecture but random weights and failed in the texture synthesis. And they claim that the learned filters are crucial for texture generation.

\paragraph{Inverting deep representations.}

\begin{figure*}[htbp]
\centering
	\vspace{-1em}
	\includegraphics[width=5in]{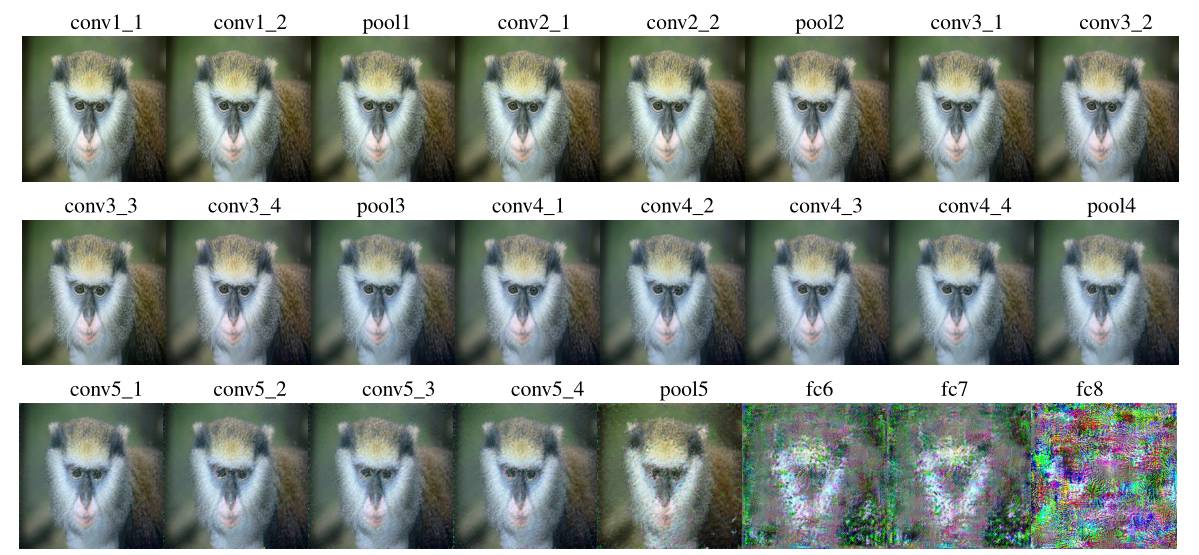}%
	\vspace{-1em}
	\caption{\textbf{Reconstructions of the monkey from each layer of the random weight CNN, ranVGG.} The monkey image is well reconstructed from activations of any of the 16 convolutional layers after the rectifier and the 5 average pooling layers, and we could still see the rough contours from the first two fully connected layers.}% The figure is best viewed in color/screen. }%The last layer fc8 rarely shows the contours and hence is ignored here. }
	\label{fig:MonkeySingleLayer} %% label for entire figure
\end{figure*}

We select several source images from the ILSVRC 2012 challenge~\cite{ILSVRC} validation data as examples for the inversion task,
and choose a monkey image as the reference image to build the stacked \emph{ranVGG}\footnote{Note that using other image as the reference image also returns similar results.}.
The reconstruction of the monkey from each layer of the \emph{ranVGG} is as shown in Figure \ref{fig:MonkeySingleLayer}.
All convolutional and pooling layers retain photographically accurate information about the image,
and the representations reveal the invariance even for the first two fully connected layers.% with different degrees of geometric and photometric invariance.
%obtain insights into the invariances captured by the representation.

\begin{figure*}[htbp]
    \centering
	\vspace{-0.5em}
	\includegraphics[width=3.5in]{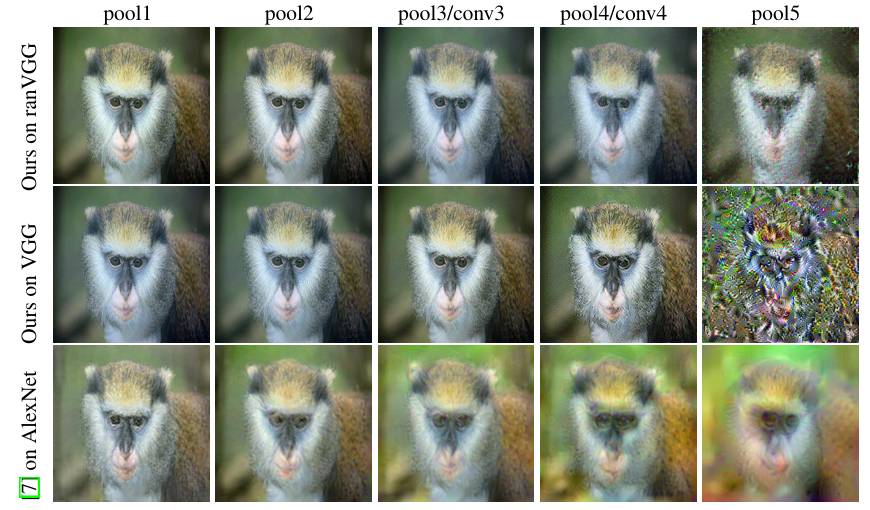}
	\vspace{-0.5em}
	\caption{\textbf{Reconstructions from layers of ranVGG (top) and the pretrained VGG (middle) and \cite{Mahendran2015CVPR} (bottom).} As AlexNet only contains 3 pooling layers, we compare their results on conv3 and conv4 with ours on pool3 and pool4. Our method on ranVGG demonstrates a higher perceptive quality, especially on the higher layers. Note that VGG is much deeper than AlexNet even when we compare on the same pooling layer.} %, even of we double the gradient descent iterations.} %The figure is best viewed in color/screen.} stronger generative power
	\label{fig:EdisCompareMonkey} %% label for entire figure
\end{figure*}

\begin{figure*}[htbp]
    \centering
    \vspace{-2em}
    \includegraphics[width=4in]{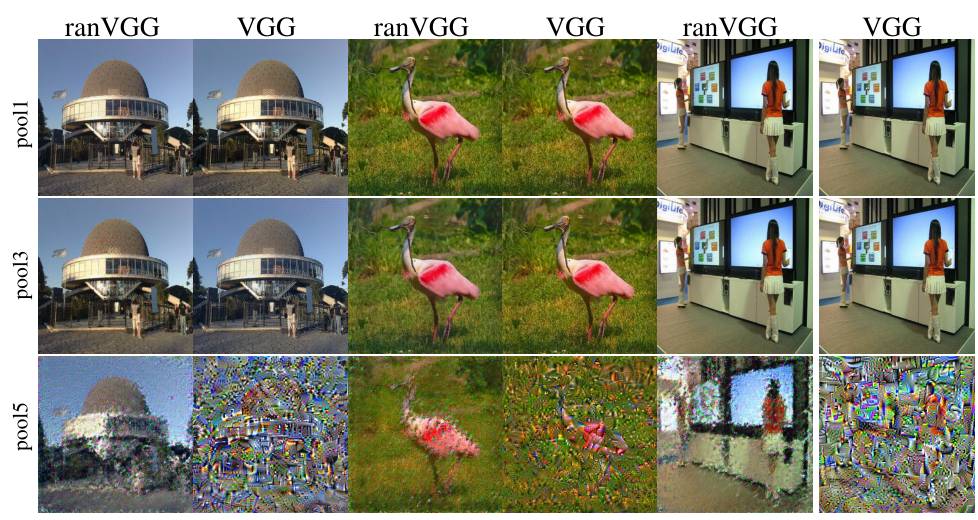}
	\vspace{-0.5em}
	\caption{\textbf{Reconstructions from different pooling layers of the untrained ranVGG and the pretrained VGG.}  ranVGG demonstrates a higher perceptive quality, especially on the higher layers. The pretrained VGG could rarely reconstruct even the contours from representations of the fifth pooling layer.} %, even of we double the gradicent descent iterations.} %The figure is best viewed in color/screen.} stronger generative power
	\label{fig:EdisMultiPooling} %% label for entire figure
\end{figure*}

As compared with the inverting technique proposed by Mahendran et al.~\cite{Mahendran2015CVPR},
we only consider the Euclidean loss over the activations and ignore the regularizer they used to capture the natural image prior.
Our \emph{ranVGG} contains 19 layers of random weights (16 convolutional layers and 3 fully connected layers), plus 5 pooling layers.
Mahendran et al. use a reference network AlexNet~\cite{Alex2012NIPS} which contains 8 layers of trained weights (5 convolutional layers and 3 fully connected layers), plus 3 pooling layers. Figure \ref{fig:EdisCompareMonkey} shows that we reach higher perceptive reconstructions\footnote{All images are best viewed in color/screen.}.
The reason may lie in the fact that the VGG architecture uses filters with a small receptive field of $3 \times 3$ and we adopt average pooling.
Though shallower than VGG, their reference network, AlexNet, adopts larger filters
%($11 \times 11$, $5 \times 5$ for the first two convolutional layers, and $3 \times 3$ for the following 3 convolutional layers)
 and uses maximum pooling, which makes it harder to get images well inverted and easily leads to spikes. That's why they used regularizers to polish the reconstructed image.
Figure \ref{fig:EdisMultiPooling} shows more examples (house, falmingo, girl) of reconstructions obtained by our method on ranVGG and on the well trained VGG.

\begin{figure}[htbp]
\centering
\includegraphics[width=2.5in]{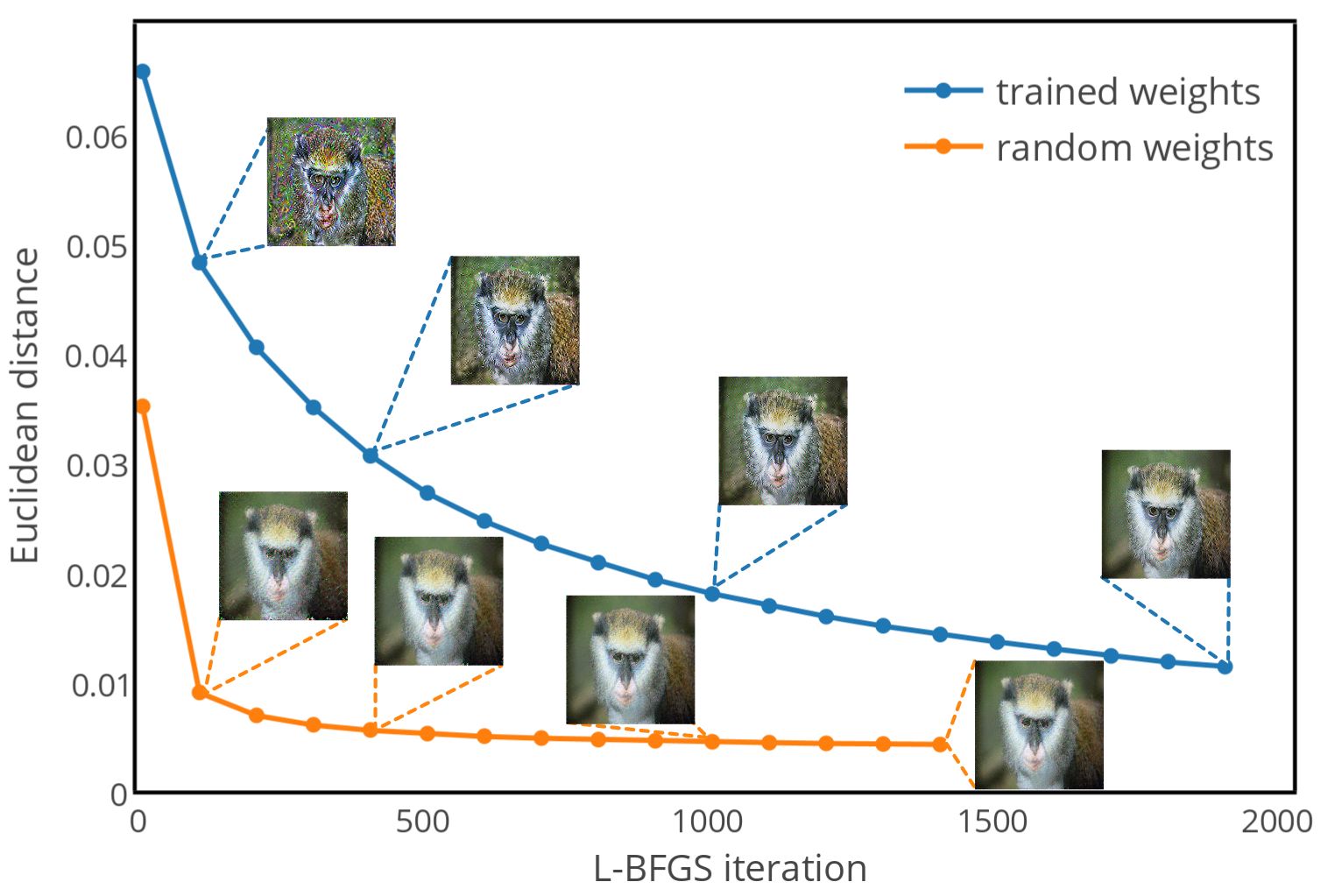}\hfill
\includegraphics[width=2.5in]{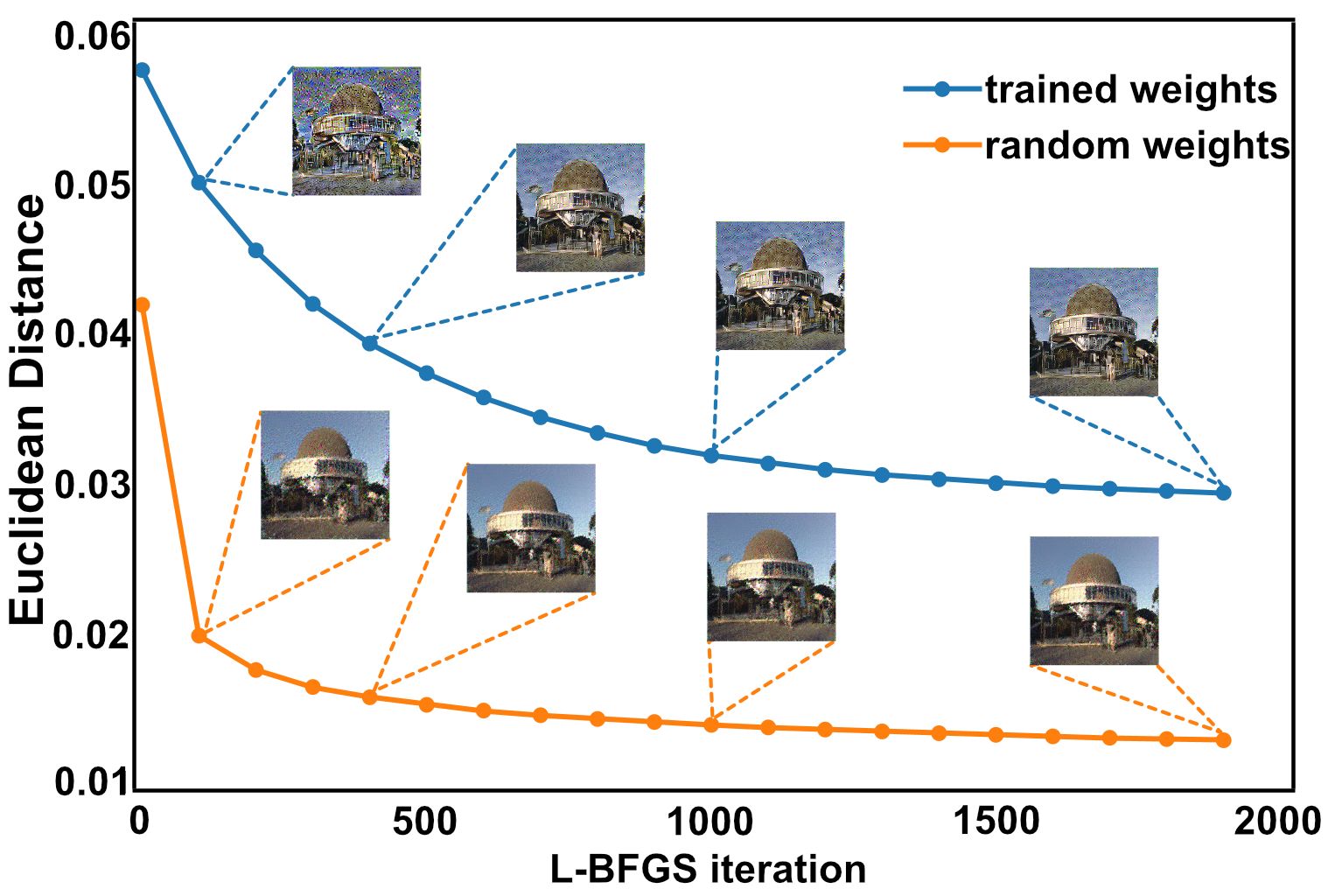}
	\vspace{-0.5em}
\caption{Reconstruction qualities of conv5\_1 during the gradient descent iterations, show comparison of our method on ranVGG and VGG. Illustrations from the 10th L-BFGS iteration. The process on the untrained ranVGG converges quickly and yields higher quality results within 100 iterations as compared with the final results after 2000 iterations on the pretrained VGG.}
\label{Fig:EdisIteration}
\end{figure}

Figure \ref{Fig:EdisIteration} shows the convergence of the loss (average Euclidean distance) along the gradient descent iterations for two example images, monkey and house.
The reconstruction converges much quicker on ranVGG and yields higher perceptual quality results. Note that the reconstruction on VGG remains the same even if we double the iteration limits to 4000 iterations.

\begin{figure}[htbp]
\centering
	\begin{minipage}[b]{0.63\textwidth}
		\centering
		\includegraphics[width=3.5in]{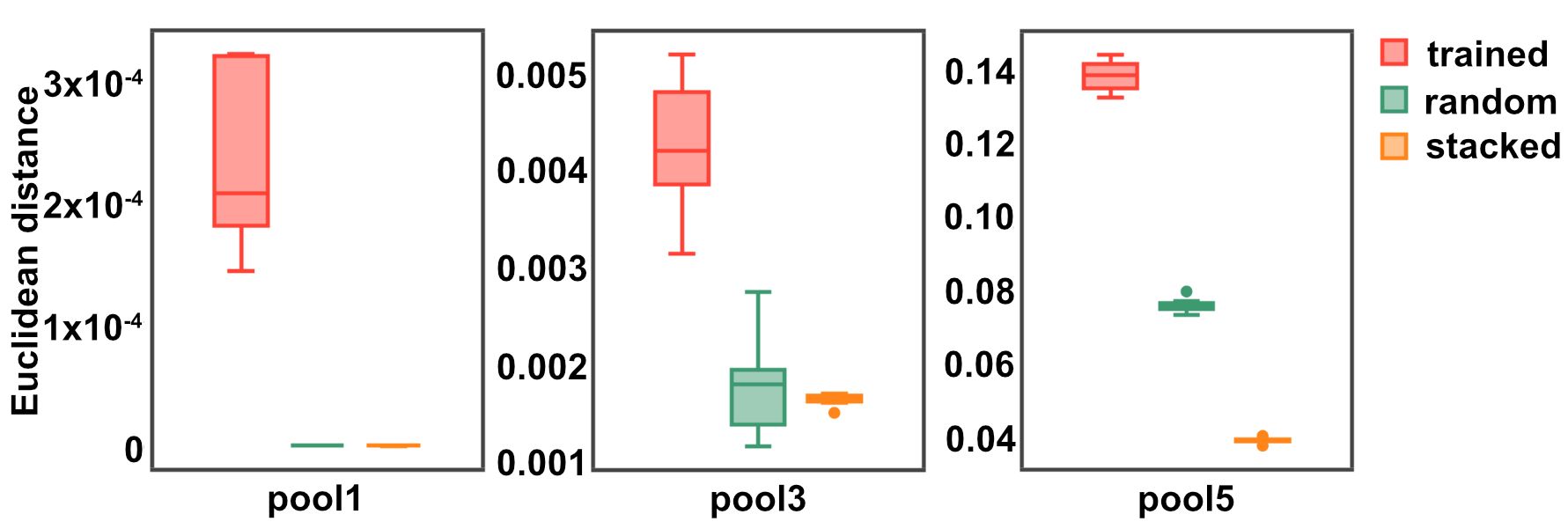}
	    \vspace{-0.5em}
		\caption{Variations in samples on the girl image, with maximum, minimum, mean and quartiles.}
    	\label{fig:girl}
	\end{minipage} \hfill
	\begin{minipage}[b]{0.33\textwidth}
		\includegraphics[width=1.8in]{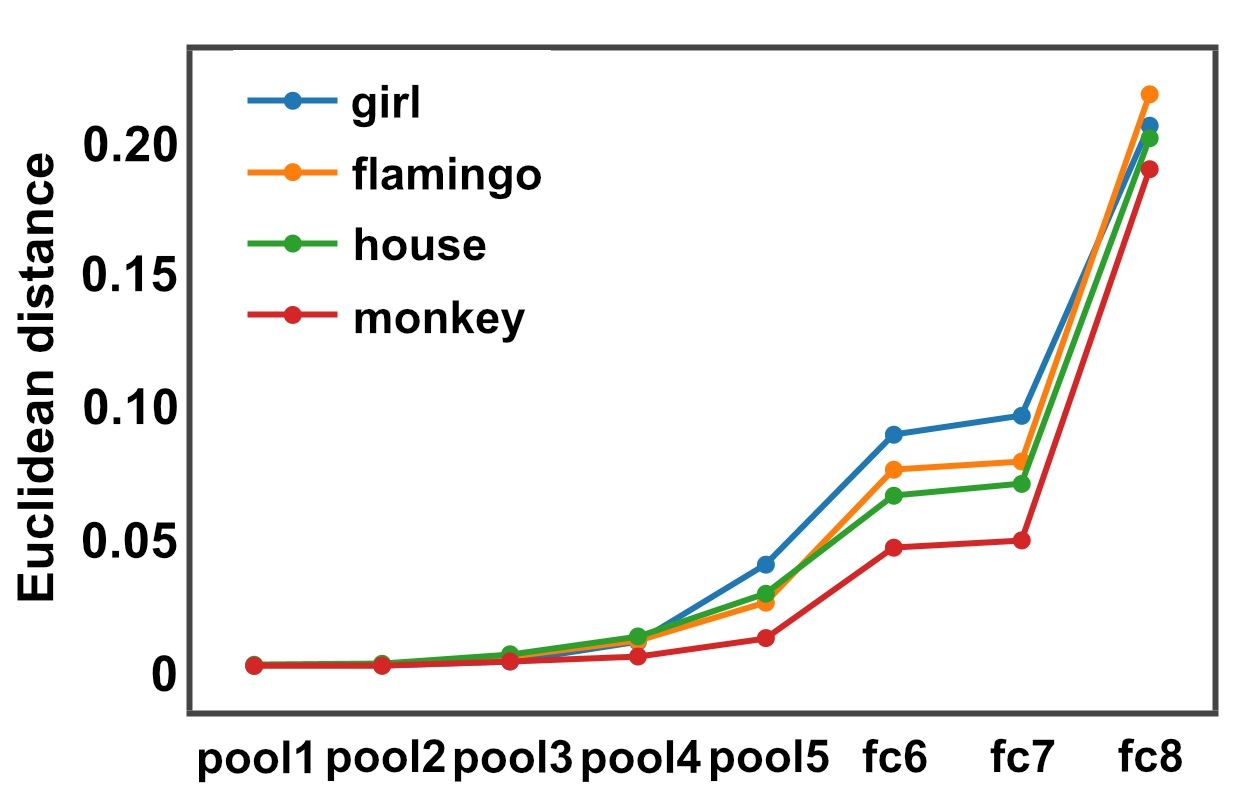}
    	\vspace{-0.5em}
	    \caption{Mean Euclidean distances on ranVGG.}
        \label{fig:meanEdisCurve}
	\end{minipage}
\end{figure}

Figure \ref{fig:girl} shows the variations on one example image (the girl image at Figure \ref{fig:EdisMultiPooling}).
As compared with the VGG with purely random weights, ranVGG, the VGG with stacked random weights, exhibits lower variations and lower reconstruction distances.
As compared with the trained VGG, both stacked ranVGG and VGG with purely random weights exhibit lower reconstruction distance with lower variations.
ranVGG demonstrates a more stable and high performance for the inversion task.

%\begin{figure}
%\centering
%\includegraphics[width=5in]{./figure/Edis/girl_combine.jpg}
%\caption{\textbf{Variations on the girl image.} Explain.}
%\label{Fig:girlsEdis}
%\end{figure}

%\begin{figure}
%\centering
%\includegraphics[width=3.5in]{./figure/Edis/layer_curve.png}
%\caption{Mean distance on the pooling layers.}
%\label{Fig:meanEdis}
%\end{figure}

% \vspace{-2mm}
%\begin{figure}
%\centering
%\includegraphics[width=3in]{./figure/Edis/monkeySingleLayer/monkey_conv1_1.png}
%\caption{Conv1\_1 of Monkey image.}
%\label{Fig:MonkeyConv11}
%\end{figure}

%\begin{figure}
%\centering
%\includegraphics[width=1.2in]{./figure/Edis/monkeySingleLayer/monkey_conv1_1.png}
%\includegraphics[width=1.2in]{./figure/Edis/monkeySingleLayer/monkey_conv1_2.png}
%\caption{Conv1\_1 of Monkey image.}
%\label{Fig:MonkeyConv11}
%\end{figure}

\begin{figure*}[htbp]
    \centering
    \includegraphics[width=5in]{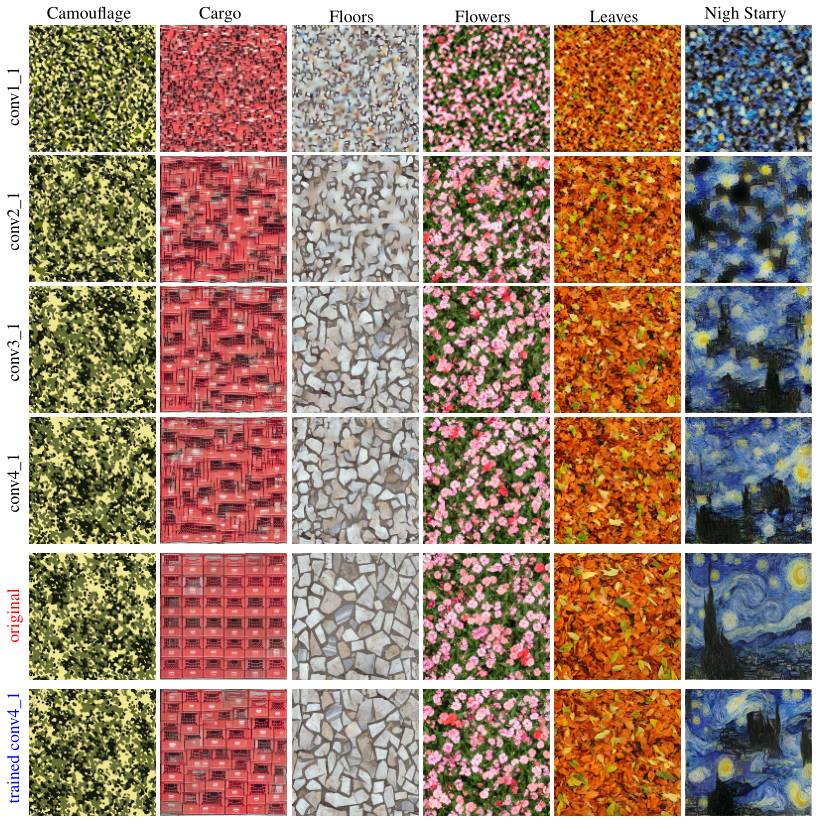}
	\vspace{-0.5em}
	\caption{\textbf{Generated textures using random weights.} Each row corresponds to a different processing stage in the random weight network ranVGG.
Considering only the lowest layer, conv1\_1, the synthesised textures are of lowest granularity, showing very local structure.
 Increasing the number of layers on which we match the texture representation (conv1\_1 plus conv2\_1 for the second row, etc),
  we have higher organizations of the previous local structure. The third row and the forth row well synthesis the textures of the original images.
  %The fifth column illustrates the textures of the artwork of starry night by Vincent van Gogh, 1889, which has global structure as compared with the previous four columns.
  The lowest row corresponds to the result of using the trained VGG to match the texture representation from conv1\_1, conv2\_1 conv3\_1 and conv4\_1.}%Matching representations of the five convolutional layers conv1\_1, conv2\_1, to conv5\_1 yields the global structure (fifth row).	 } %The figure is best viewed in color/screen.}
	\label{fig:texture} %% label for entire figure
\end{figure*}

\paragraph{Texture synthesis.}
Figure \ref{fig:texture} shows the textures synthesized by our model on ranVGG
for several natural texture images (fifth row) selected from a texture website\footnote{http://www.textures.com/} and an artwork named \emph{Night Starry} by Vincent van Gohn 1989.
Each row of images was generated using an increasing number of convolutional layers to constrain the gradient descent.
conv1\_1 for the first row, conv1\_1 and conv2\_1 for the second row, etc (the labels at each row indicate the top-most layer included).
The joint matching of conv1\_1, conv2\_1, and con3\_1 (third row) already exhibits high quality texture representations.
Adding one more layer of conv4\_1 (forth row) could slightly improve the natural textures.
%By comparison, results of our model on the trained VGG using four convolutional layers up to conv4\_1 are as shown at the bottom row.
By comparison, results of Gatys et al.\cite{Gatys2015texture} on the trained VGG using four convolutional layers up to conv4\_1 are as shown at the bottom row.

Our experiments show that with suitable weighted factors, calculated automatically by our method, % described in the previous section,
ranVGG could synthesize complex natural textures that are almost indistinguishable with the original texture and the synthesized texture on the trained VGG.
Trained VGG generates slightly better textures on neatly arranged original textures (cargo at the second column).

\begin{figure}[htbp!]
	\subfigure{
		\begin{minipage}[b]{0.014\textwidth}
			\rotatebox{90}{~~\scriptsize{Original}~~~~~~	} \\
			\rotatebox{90}{~~\scriptsize{Ours on ranVGG}~~~~~~} \\
			\rotatebox{90}{~~~~~\scriptsize{\cite{Gatys2015Style} on VGG}~~~~~}
		\end{minipage}}
	\subfigure{%[pool1]
		\begin{minipage}[b]{0.98\textwidth}
        ~~~~~~~~~~~\scriptsize{Night Starry}~~~~~~~
        ~~~~~~~~~~~~~~~~~~~~~~~~~\scriptsize{Der Schrei}~~~~~~~~~~~~~~
        ~~~~~~~~~~~~~~~~~~\scriptsize{Photograph}~~~~~~~
        ~~~~~~~~~~~~~\scriptsize{Picasso}~~~~\scriptsize{Woman with a Hat}~~~~
        ~~~~\scriptsize{Meadow with Poplars}~\\
        \centering
        \includegraphics[width=5.5in]{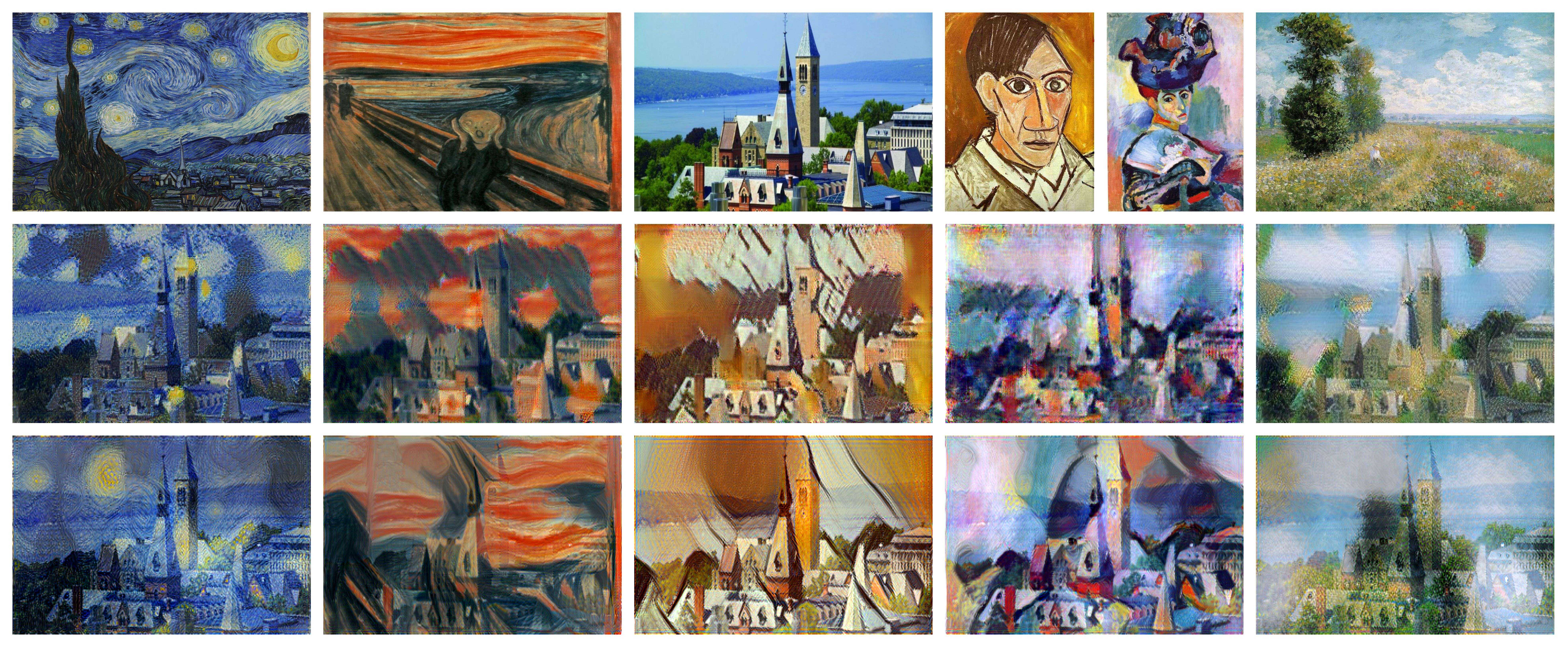}
        \end{minipage}}
        \caption{
					\textbf{%Images that combine the content of a photograph with the style of several well-known artworks.
                             Artistic style images of ours on the untrained ranVGG (medium row) and
						    of Gatys et al.\cite{Gatys2015texture} on the pretrained VGG (bottom row)}.
We select a university image (first row, center) and several well-known artworks for the style (first row, others images).
The third column under the photograph are for the Picasso.
We obtain similar quality results as compared with Gatys et al.\cite{Gatys2015Style}.
				}
\vspace{-2em}
\label{Fig:Cornell}
\end{figure}

\begin{figure}[htbp!]
  \vspace{-4em}
%	\subfigure{
%		\begin{minipage}[b]{0.02\textwidth}
%			\rotatebox{90}{~~~~~~~~~~~~\scriptsize{picture 1}~~~~~~} \\
%			\rotatebox{90}{~~~~~~~~~~~~~~\scriptsize{picture 2}~~~~~~~~~~~}\\
%			\rotatebox{90}{~~~~~~~~~~~~\scriptsize{picture 3}~~~~~~}
%		\end{minipage}}
	\subfigure{%[pool1]
		\begin{minipage}[b]{0.98\textwidth}
        ~~~~~~\scriptsize{Chinese painting}~~~~~~~~~~~~~~~~~~~~~
        ~~~~~~~~~~~~~~~~~~~~~~~~~~~~~~~\scriptsize{Photograph}~~~~~~~~~~~~~~~~~~~
        ~~~~~~~~~~~~~~~~~~~~~~~~~~~~~~~~~~~~~\scriptsize{Created image}~~~\\
        \centering
        \includegraphics[width=5.5in]{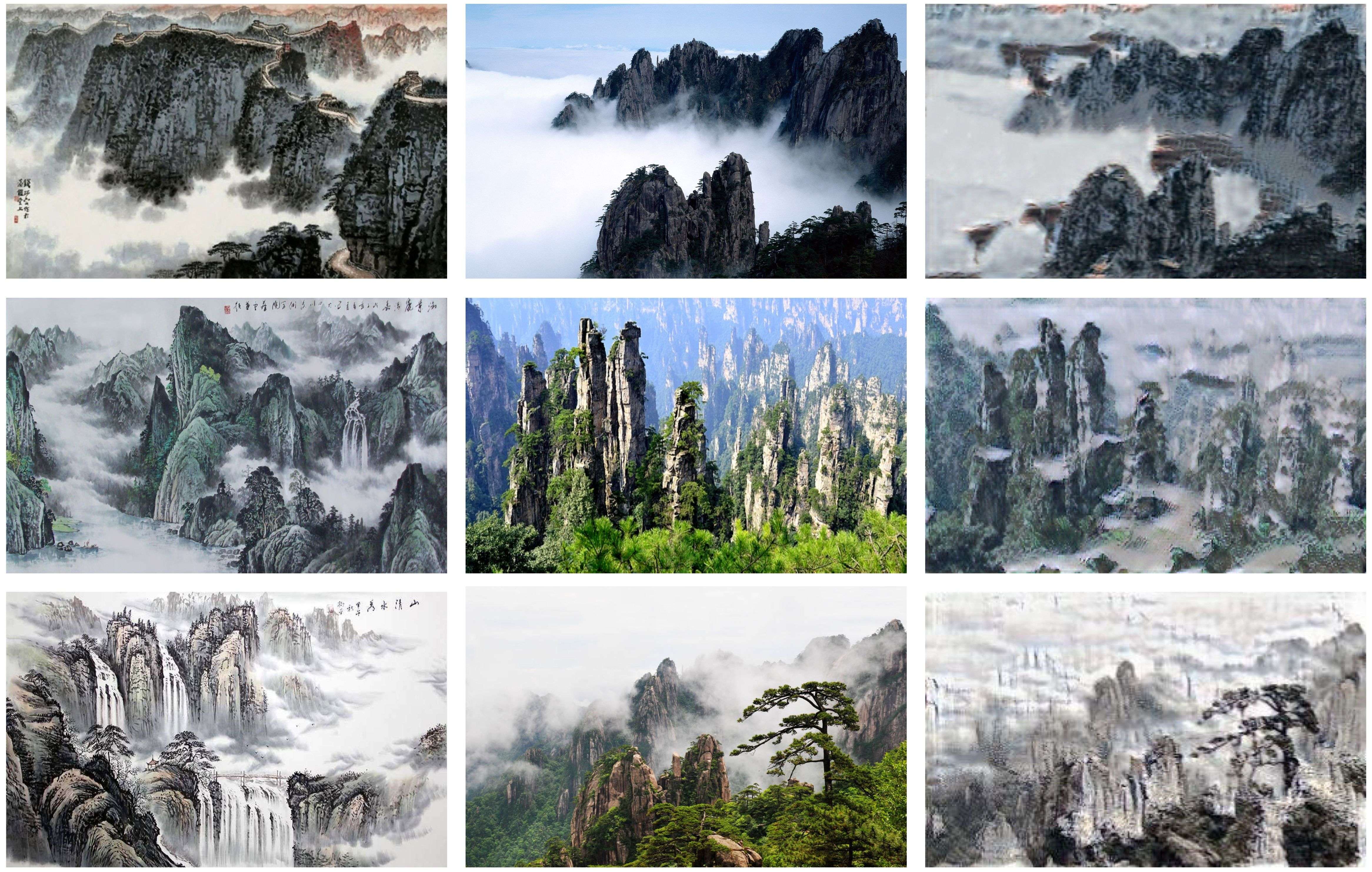}
        \end{minipage}}
        \vspace{-1em}
\caption{
					\textbf{Style transfer of Chinese paintings on the untrained ranVGG.}
					We select several Chinese paintings for the style (first column),
					including \emph{The Great Wall} by Songyan Qian 1975, a painting of anonymous author
					%\textbf{B} \emph{},
					and \emph{Beautiful landscape} by Ping Yang.
					We select the mountain photographs (second column) as the content images.
					The created images performed on the untrained ranVGG are shown in the third column,
                    which seem to have learned how to paint the rocks and clouds from paintings of the first column and transfer the style to the content to ``draw'' Chinese landscape paintings.
 				}
\label{Fig:ChinesePainting}
\end{figure}

\paragraph{Style transfer.}
%We select conv1\_1,conv2\_1, conv3\_1, conv4\_1, conv5\_1 as the style representation layers, and conv2\_2 as the content representation layer
% for our artistic image generation on ranVGG.
We select conv2\_2 as the content layer, and use the combination of conv1\_1,  conv2\_1, ..., conv5\_1 as the style.
We set the ratio of $\alpha :\beta : \gamma = 100 : 1 : 1000$ in the experiments.
We first compare our style transfer results with the prior work of Gatys et al.\cite{Gatys2015Style} on several well-known artworks for the style:
\emph{Night Starry} by Vincent van Gohn 1989, \emph{Der Schrei} by Edward Munch 1893,
\emph{Picasso} by Pablo Picasso 1907, \emph{Woman with a Hat} by Henri Matisse 1905, \emph{Meadow with Poplars} by Claude Monet 1875.
As shown in Figure \ref{Fig:Cornell}, the second row, by recasting the content of a university image in the style of the five artworks,
we obtain different artistic images based on the untrained ranVGG (second row).
Our results are comparable to their work \cite{Gatys2015Style} on the pretrained VGG (third row), and are in the same order of magnitude. They have slightly smoother lines and textures which may attributed to the training.
%The reconstructed images and the photograph are in the same aspect ratio.
We further try the content and style combination on some Chinese paintings and scenery photographs, as shown in Figure \ref{Fig:ChinesePainting},
and create high perceptual artistic Chinese paintings that well combine the style of the painting and the content of the sceneries.

\vspace{-0.5em}
\section{Discussion}

Our work offers a testable hypothesis about the representation of image appearance based only on the network structure.
The success on the untrained, random weight networks on deep visualization raises several fundamental questions in the area of deep learning.
Researchers have developed many visualization techniques to understand the representation of well trained deep networks. However, if we could do the same or similar visualization using an untrained network, then the understanding is not for the training but for the network architecture.
What is the difference of a trained network and a random weight network with the same architecture, and how could we explore the difference?
What else could one do using the generative power of untrained, random weight networks?
Explore other visualization tasks in computer vision developed on the well-trained network, such as image morphing~\cite{Traversal2015}, would be a promising aspect.

%%%help the selection of architecture and hyperparameters
%Training deep neural networks not only requires a long time but also significant high performance computational resources.
%% on the network architecture.
%The VGG network, which contains 11-19 weight layers depending on the typical architecture~\cite{Simonyan14VGG},
% takes 2 to 3 weeks on a system equipped with 4 NVIDIA Titan Black GPUs for training a single net.
%The residual network ResNet, which achieved state-of-the-art results in image classification and detection in 2015~\cite{He2015Residual}, takes
% 3.5 days for the 18-layer model and 14 days for the 101-layer model using 4 NVIDIA Kepler GPU.\footnote{http://torch.ch/blog/2016/02/04/resnets.html}
%Could we evaluate a network structure without taking a long time to train it?
%In future work, we will address the above issue by utilizing the untrained network to attempt to compare networks quickly without having to train them.

\section*{Acknowledgments}
This research work was supported by US Army Research Office(W911NF-14-1-0477), National Science Foundation of China(61472147) and National Science Foundation of Hubei Province(2015CFB566).

%\newpage
\small
\bibliographystyle{unsrt}%abbrv}%{alpha}
\bibliography{references}

%\section*{References}
%[1] Alexander, J.A.\ \& Mozer, M.C.\ (1995) Template-based algorithms
%for connectionist rule extraction. In G.\ Tesauro, D.S.\ Touretzky and
%T.K.\ Leen (eds.), {\it Advances in Neural Information Processing
%  Systems 7}, pp.\ 609--616. Cambridge, MA: MIT Press.
%
%[2] Bower, J.M.\ \& Beeman, D.\ (1995) {\it The Book of GENESIS:
%  Exploring Realistic Neural Models with the GEneral NEural SImulation
%  System.}  New York: TELOS/Springer--Verlag.
%
%[3] Hasselmo, M.E., Schnell, E.\ \& Barkai, E.\ (1995) Dynamics of
%learning and recall at excitatory recurrent synapses and cholinergic
%modulation in rat hippocampal region CA3. {\it Journal of
%  Neuroscience} {\bf 15}(7):5249-5262.

%\newpage
%\normalsize
%\input{080Appendix}

\end{document}